\documentclass[letterpaper]{article} 
\usepackage{aaai24}  
\usepackage{times}  
\usepackage{helvet}  
\usepackage{courier}  
\usepackage[hyphens]{url}  
\usepackage{graphicx} 
\usepackage{natbib}  
\usepackage{caption} 
\usepackage{algorithm}

\usepackage{newfloat}
\usepackage{listings}
\DeclareCaptionStyle{ruled}{labelfont=normalfont,labelsep=colon,strut=off} 
\floatstyle{ruled}
\newfloat{listing}{tb}{lst}{}
\floatname{listing}{Listing}
\title{Cluster-based Sampling in Hindsight Experience Replay for Robotic Tasks (Student Abstract)}
\author {
    Taeyoung Kim,
    Dongsoo Har
}
\affiliations {
    Korea Advanced Institute of Science and Technology, Daejeon 34051, Korea\\
    ngng9957@kaist.ac.kr, dshar@kaist.ac.kr
}

\begin{document}

\maketitle

\begin{abstract}
In multi-goal reinforcement learning with a sparse binary reward, training agents is particularly challenging, due to a lack of successful experiences. To solve this problem, hindsight experience replay (HER) generates successful experiences even from unsuccessful ones. However, generating successful experiences from uniformly sampled ones is not an efficient process. In this paper, the impact of exploiting the property of achieved goals in generating successful experiences is investigated and a novel cluster-based sampling strategy is proposed. The proposed sampling strategy groups episodes with different achieved goals by using a cluster model and samples experiences in the manner of HER to create the training batch. The proposed method is validated by experiments with three robotic control tasks of the OpenAI Gym. The results of experiments demonstrate that the proposed method is substantially sample efficient and achieves better performance than baseline approaches.
\end{abstract}

\section{Introduction}
Reinforcement learning (RL) is a powerful framework for training an agent to take sequential actions to complete a task. The RL framework for learning multiple goals is called multi-goal RL \cite{mgrl}. In both RL frameworks, the training data is a set of experiences, which are obtained through exploration and stored in a replay buffer. The multi-goal RL environments cause a lack of successful experiences in the replay buffer due to sparse binary rewards, leading to challenging training of the agent. To mitigate this problem, hindsight experience replay (HER) was proposed in \cite{her}. HER improves sampling efficiency by generating successful experiences, named hindsight experiences, from experiences in the replay buffer that contains both unsuccessful and successful ones. However, generating hindsight experiences from successful experiences is less efficient in sampling compared to generating from failed ones.

In this paper, it is found that the efficiency of HER can be increased if experiences can be sampled in consideration of the property of achieved goals. From this viewpoint, a cluster-based sampling strategy is proposed to improve the sampling efficiency of HER.

\section{Proposed Method}
\begin{figure}[h]
\centering
\includegraphics[width=0.47\textwidth]{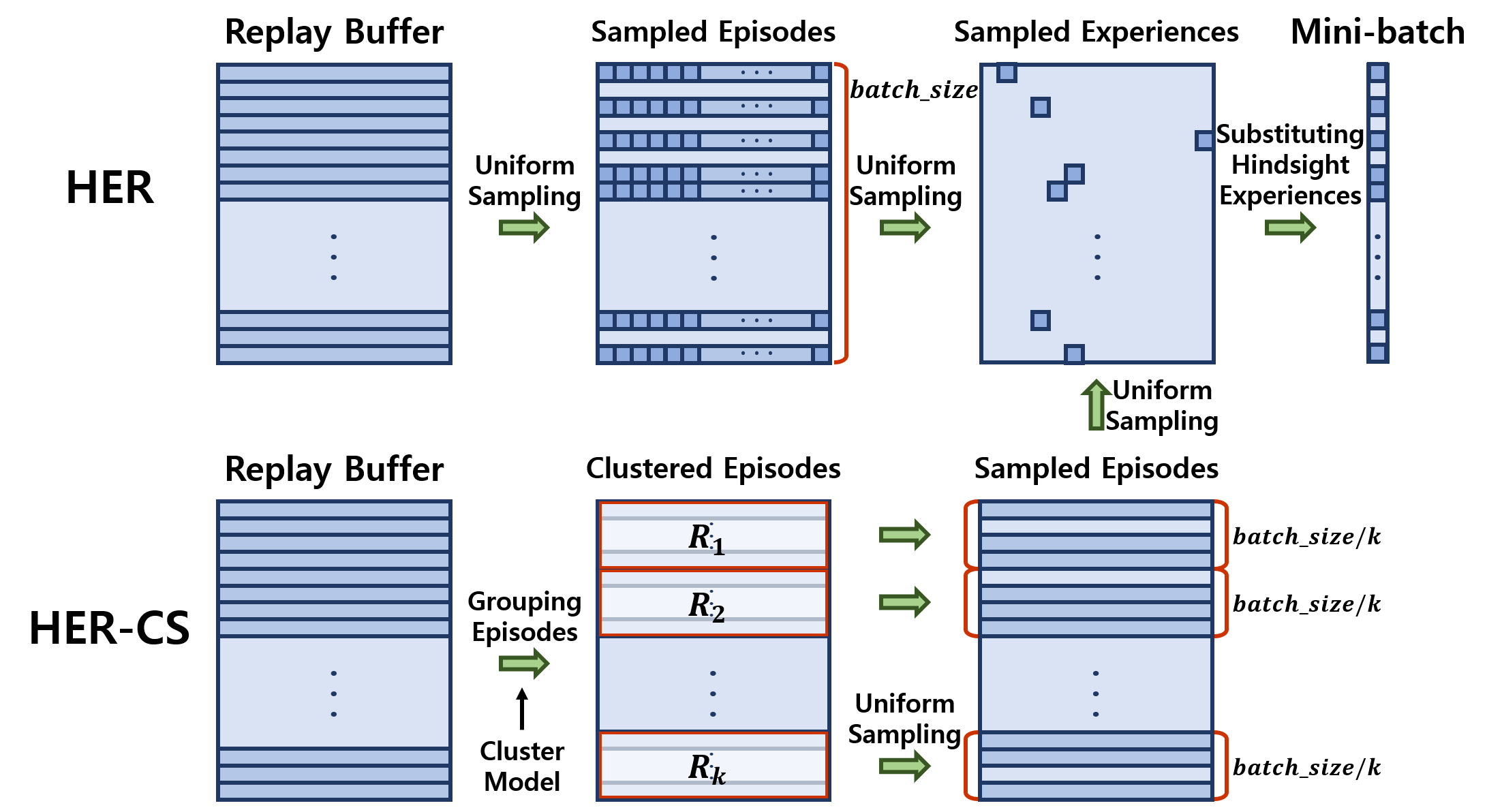}
\caption{Frameworks of hindsight experience replay (HER) and HER with cluster-based sampling (HER-CS).}
\label{frameworks}
\end{figure}

In this section, a novel cluster-based sampling strategy for HER is presented. The proposed sampling strategy involves the incorporation of a grouping process into the original HER framework. The framework of HER includes three processes of uniform samplings. As shown in the upper part of Fig~\ref{frameworks}, the first process is for sampling episodes from the replay buffer, the second one is for sampling one experience from each sampled episode, and the last one is to sample experiences to be substituted by hindsight experiences among sampled experiences in the second process. The first process of HER can be designed in a way that ``hard episodes'' are more likely sampled instead of using uniform sampling, thereby enhancing the effectiveness of HER. The ``hard episode'' refers to an episode whose last achieved goal is hard to achieve by the current RL policy. To realize this approach, HER with cluster-based sampling (HER-CS) replaces the first sampling process with two procedures: grouping of episodes using a cluster model and sampling the episodes uniformly from grouped episodes, as depicted in the lower part of Fig~\ref{frameworks}.

The grouping procedure is based on a cluster model. To group the episodes with consideration of ``hard episodes'', the cluster model is generated by applying a clustering algorithm on failed goals (FG) that are the original goals not achieved. The cluster model assigns cluster indices to episodes in the replay buffer based on their last achieved goals. According to the cluster indices, the episodes are grouped into clustered buffers. The clustered buffers $R_i$ are a subset of the replay buffer and the number thereof is equal to the number of the groups $k$. From each clustered buffer, ($batch\_size/k$) episodes are uniformly sampled to form an episode batch containing $batch\_size$ episodes. With the episode batch, the second and third uniform samplings are performed in the manner of HER.

Throughout the exploration of the RL model, the FGs are stored in the failed goal buffer (FGB). As the FGs accumulate, the cluster model is periodically updated, because continuing to use the old cluster model generated with old FGs can interfere with the training of the RL model. When the FGB is filled with entirely new FGs, the cluster model is updated. Once the cluster model is updated, the cluster model provides the cluster index for each episode in the replay buffer. For the episodes stored after the update, the cluster index is given individually by the cluster model. The important variables in this periodical update are the number of FGs used to update and the frequency of the update. Ablation studies of these important variables, including preliminary experiments, are provided in the supplemental file.

\section{Experimental Results}
In this section, we present the results of two experiments conducted by using the proposed method alongside two baseline algorithms: HER and Energy-based hindsight experience prioritization (HER-EBP) \cite{ebp}. HER-EBP is HER with a different kind of sampling strategy. Experiments are conducted for three tasks in multi-goal environments discussed in \cite{mgrl}. The three tasks are Push, PickAndPlace, and Slide tasks. The integration of the proposed sampling strategy into HER is based on the implementation in \cite{impl}.

\begin{figure}[b!]
\centering
\includegraphics[width=0.45\textwidth]{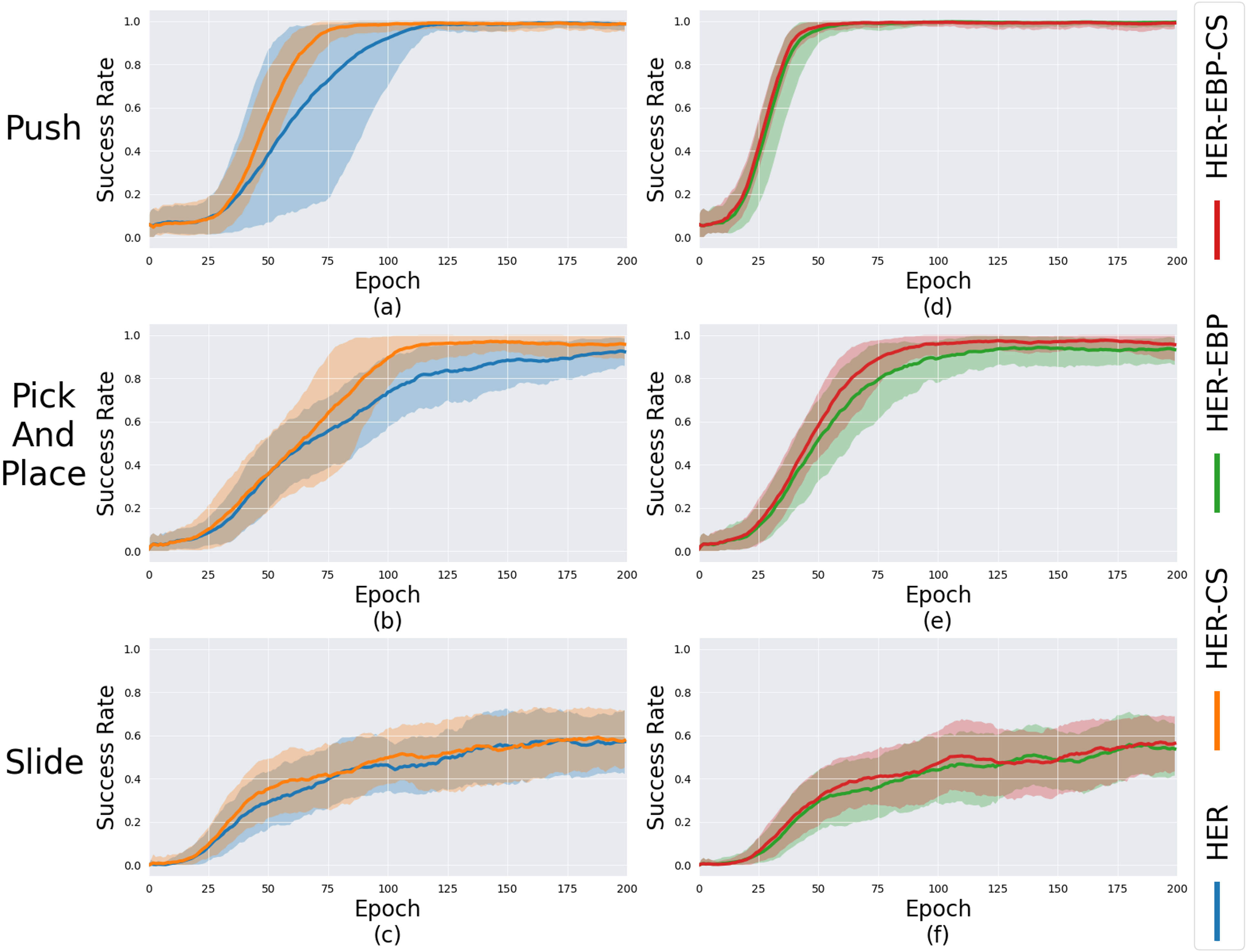}
\caption{Comparison of training performance: (a)-(c) HER vs. HER-CS; (d)-(f) HER-EBP vs. HER-EBP-CS.}
\label{results}
\end{figure}

The performance of the proposed method is evaluated in terms of the success rate. After each epoch of 200 epochs for training, the success rate is calculated with 20 test episodes. The sequence of training followed by evaluation is repeated with 5 random seeds. In the graphs in Fig~\ref{results}, the solid curve shows the average of five success rates for each epoch and the upper and lower boundary lines of the shaded area show the minimum and maximum success rates.

The left side of Fig~\ref{results} compares the performances of HER and HER-CS for the three tasks. The right side compares the performances of HER-EBP and HER-EBP with the proposed sampling strategy (HER-EBP-CS) for the same tasks. Fig~\ref{results}(a), (b), (d), and (e) illustrate that the proposed method enhances the performance of both HER and HER-EBP. It can be seen that the convergence speeds in the Push task and the maximum success rates in the PickAndPlace task are improved. As depicted in Fig~\ref{results}(c) and (f), although there are marginal performance enhancements, all four algorithms achieve comparable success rates in the Slide task, indicating the inherent difficulty of this particular task.

\section{Discussion \& Conclusion}
In this paper, the effect of leveraging the property of achieved goals within HER process on performance enhancements in the baselines is discussed with the proposed method exploiting cluster-based sampling. As shown in the graphs for the Push and PickAndPlace task, the proposed method reduces the width of the shaded area, which represents the variation in success rates across random seeds. This indicates that the proposed method can enhance training stability regardless of the initial random seed used. Furthermore, integrating the proposed method with HER-EBP, which employs a different sampling strategy for HER, demonstrates performance improvements. This suggests the upside potential that can be realized by combining the proposed method with other sampling strategies.

\section{Acknowledgments}
This work was supported by the Institute for Information communications Technology Promotion (IITP) grant funded by the Korean government (MSIT) (No.2020-0-00440, Development of Artificial Intelligence Technology that continuously improves itself as the situation changes in the real world).

\bigskip

\bibliography{aaai24}

\end{document}